\pdfoutput=1

\documentclass[11pt]{article}

\usepackage{acl}

\usepackage{times}
\usepackage{latexsym}

\usepackage[T1]{fontenc}
\usepackage[utf8]{inputenc}

\usepackage{microtype}

\usepackage{makecell}
\usepackage{multirow}
\usepackage{diagbox}
\usepackage{pgfplots}
\usepackage{color}
\usepackage{mathtools}
\usepackage{amssymb}
\usepackage{graphicx}
\usepackage{subfigure} 
\usepackage{booktabs}
\usepackage{algorithm}
\usepackage{algorithmic}
\usepackage{hyperref}
\usepackage{soul}
\usepackage{xspace}
\newcommand{\method}{\textsc{TF-Attack}\xspace}
\title{TF-Attack: Transferable and Fast Adversarial Attacks\\ on Large Language Models}

\author{Zelin Li\textsuperscript{\rm $1$}, Kehai Chen\textsuperscript{\rm $1$}, Lemao Liu, Xuefeng Bai\textsuperscript{\rm $1$}, Mingming Yang, Yang Xiang\textsuperscript{\rm $2$} and Min Zhang\textsuperscript{\rm $1$} \\
$\textsuperscript{\rm $1$}$School of Computer Science and Technology, Harbin Institute of Technology, Shenzhen, China \\
$\textsuperscript{\rm $2$}$Pengcheng Laboratory, Shenzhen, China \\
\texttt{22S151142@stu.hit.edu.cn}, \texttt{\{chenkehai,baixuefeng\}@hit.edu.cn}, \\
\texttt{\{lemaoliu,shanemmyang\}@gmail.com}, \\ \texttt{xiangy@pcl.ac.cn}, \texttt{zhangmin2021@hitsz.edu.cn}
}

\begin{document}

\maketitle

\begin{abstract}
With the great advancements in large language models (LLMs), \textit{adversarial attacks} against LLMs have recently attracted increasing attention.
We found that pre-existing adversarial attack methodologies exhibit limited transferability and are notably inefficient, particularly when applied to LLMs.
In this paper, we analyze the core mechanisms of previous predominant adversarial attack methods, revealing that 1) the distributions of importance score differ markedly among victim models, restricting the transferability; 2) the sequential attack processes induces substantial time overheads.
Based on the above two insights, we introduce a new scheme, named \method, for \textbf{T}ransferable and \textbf{F}ast adversarial attacks on LLMs. \method employs an external LLM as a third-party overseer rather than the victim model to identify critical units within sentences.
Moreover, \method introduces the concept of \textit{Importance Level}, which allows for parallel substitutions of attacks.
We conduct extensive experiments on 6 widely adopted benchmarks, evaluating the proposed method through both automatic and human metrics.
Results show that our method consistently surpasses previous methods in transferability and delivers significant speed improvements, up to 20$\times$ faster than earlier attack strategies.


\end{abstract}

\section{Introduction}

Recently, large language models (LLMs) such as ChatGPT and LLaMA~\citep{ouyang2022training, touvronllama} have demonstrated considerable promise across a range of downstream tasks~\citep{kasneci2023chatgpt, thirunavukarasu2023large, liu2023summary}. 
Subsequently, there has been increasing attention on the task of adversarial attack~\cite{xu2023llm, yao2023llm}, which aims to generate adversarial examples that confuse or mislead LLMs. 
This task is crucial for advancing reliable and robust LLMs in the AI community, emphasizing the paramount importance of security in AI systems~\citep{Marcus2020,thiebes2021trustworthy}.


\begin{table}[t!]
    \centering
    \small
    \scalebox{.95}{
    \begin{tabular}{l@{\hspace{2pt}}cccc@{\hspace{2.2pt}}c@{\hspace{2.2pt}}}
    \toprule
         & \textbf{CNN} & \textbf{LSTM} & \textbf{BERT} & \textbf{LLaMA} &\textbf{Baichuan} \\
    \midrule
    \textbf{CNN}  & \textbf{94.7} & 22.9 & 19.8  & 21.7  & 21.2  \\
    \textbf{LSTM} & 17.3 &  \textbf{94.1} & 19.4  & 22.0  & 22.1 \\
    \textbf{BERT}     & 12.6 & 14.6 & \textbf{91.0} & 16.6  & 18.3  \\
    \textbf{LLaMA}   & 12.1 & \ \ 9.5 & \ \ 8.7  & \textbf{86.1} & 16.0  \\
    \textbf{Baichuan}  & 21.8 & 24.0 & 19.6  & 27.2  & \textbf{89.3} \\
    \textbf{ChatGPT}  & 11.2 & 14.4 & 12.6  & 16.2 & 14.8  \\ 
    \midrule
    \textbf{Average}  & 14.6 & 17.0 & 18.0 & 20.7 & 18.1 \\ 
    \bottomrule
    \end{tabular}
    }
    \caption{Transferability evaluation of BERT-Attack samples on IMDB dataset. Row $i$ and column $j$ is the Attack Success Rate of samples generated from model $j$ and evaluated on model $i$. The Average result is from non-diagonal elements of each column.}
    \label{tab:bad_of_BA}
    \vspace{-0.5cm}
\end{table}

Existing predominant adversarial attack approaches on LLMs typically adhere to a two-step process: initially, they rank token importance based on the victim model, and subsequently, they replace these tokens sequentially following specific rules \citep{cer2018universal, oliva2011symss, jin2020bert}.
Despite notable successes, recent studies highlight that current methods face two substantial limitations: 1) poor transferability of the generated adversarial samples.
As depicted in Table~\ref{tab:bad_of_BA}, while adversarial samples generated by models can drastically reduce their own classification accuracy, they scarcely affect other models; 
2) significant time overhead, particularly with Large Language Models \citep{spector2023accelerating}. For instance, the time required to conduct an attack on LLaMA is 30$\times$ slower than standard inference.


To tackle the above two issues, we begin by analyzing the causes of the poor transferability and the slow speed of existing adversarial attack methods. 
Specifically, we first study the impact of the importance score, which is the core mechanism of previous methods. Comparative analysis reveals distinct importance score distributions across various victim models. 
This discrepancy largely explains why the portability of adversarial samples generated by existing methods is poor, since perturbations generated according to a specific pattern of one model do not generalize well to other models with different importance assignments.
In addition, analysis of time consumption in Figure~\ref{fig:timecost} shows that over 80$\%$ of processing time is spent on sequential word-by-word operations when attacking LLMs.

Drawing on insights from the above observations,
we propose a new scheme, named \method, for transferable and fast adversarial attacks over LLMs. 
\method follows the overall framework of BERT-Attack which generates adversarial samples by synonym replacement.
Different from BERT-Attack, \method employs an external third-party overseer such as ChatGPT to identify important units with the input sentence, thus eliminating the dependency on the victim model. 
Moreover, \method introduces the concept of \textit{Important Level}, which divides the input units into different groups based on their semantic importance. 
Specifically, we utilize the powerful abstract semantic understanding and efficient automatic extraction capabilities of ChatGPT to form an important level priority queue through human-crafted \textit{Ensemble Prompts}.
Based on this, \method can perform parallel replacement of entire words within the same priority queue, as opposed to the traditional approach of replacing them sequentially one by one, as shown in Figure~\ref{fig:framework}. 
This approach markedly reduces the time of the attacking process, thereby resulting in a significant speed improvement.
Furthermore, we employ two tricks, named \textit{Multi-Disturb} and \textit{Dynamic-Disturb}, to enhance the attack effectiveness and transferability of generated adversarial samples. 
The former involves three levels of disturbances within the same sentence, while the latter dynamically adjusts the proportions and thresholds of the three types of disturbances based on the sentence length of the input.
They significantly boost attack effectiveness and transferability, and are adaptable to other attack methods.

We conduct experiments on six widely adopted benchmarks and compare the proposed method with several state-of-the-art methods.
To verify the effectiveness of our method, we consider both automatic and human evaluation. 
Automatic evaluation shows that \method has surpassed the baseline method TextFooler, BERT-Attack and SDM-Attack.
In addition, human evaluation results demonstrate that \method maintains a comparable consistency and achieves a comparable level of language fluency that does not cause much confusion for humans.
Moreover, we compare the transferability of \method with BERT-Attack, demonstrating that \method markedly diminishes the accuracy of other models and exhibits robust migration attack capabilities.
Furthermore, the time cost of \method is significantly lower than BERT-Attack, more than 10$\times$ speedup on average stats.
Lastly, the adversarial examples generated by \method minimally impact the performance of model after adversarial training, significantly strengthening its defense against adversarial attacks.
Overall, the main contributions of this work can be summarized as follows:

\begin{itemize}
    \item We investigate the underlying causes behind the slow speed and poor effectiveness of pre-existed adversarial attacks on LLMs.
    
    \item We introduce \method, a novel approach leveraging an external LLM to identify critical units and facilitates parallelized adversarial attacks. 

    \item \method effectively enhances the transferability of generated adversarial samples and achieves a significant speedup compared to previous methods.
\end{itemize}

\section{Related Work}
\subsection{Text Adversarial Attack}
For NLP tasks, the adversarial attacks occur at various text levels including the character, word, or sentence level. Character-level attacks involve altering text by changing letters, symbols, and numbers. Word-level attacks \citep{wei2019eda} involve modifying the vocabulary with synonyms, misspellings, or specific keywords. Sentence-level attacks \cite{coulombe2018text, xie2020unsupervised} involve adding crafted sentences to disrupt the output of model. Current adversarial attacks in NLP\citep{alzantot2018generating, jin2020bert} employ substitution to generate adversarial examples through diverse strategies, such as genetic algorithms \citep{zang2020word,guo2021gradient}, greedy search \citep{sato2018interpretable,yoo2021towards}, or gradient-based methods \citep{ebrahimi2018adversarial, cheng2018towards}, are employed to identify substitution words form synonyms \citep{jin2020bert} or language models \citep{li2020bert,garg2020bae,li2021contextualized}. Recent studies have refined sampling methods, yet these approaches continue to be time-intensive, highlighting a persistent challenge in efficiency. \citep{fang2023modeling} apply reinforcement learning, showing promise on small models but facing challenges on LLMs \citep{ji2024feature, zhong2024understanding, jiang2024survey} due to lengthy iterations, limiting large-scale adversarial samples.

\subsection{Sample Transferability}
Evaluating text adversarial attacks heavily depends on sample transferability, assessing the performance of attack samples across diverse environments and models to measure their broad applicability and consistency. In experiments~\cite{qi2021mind}, adversarial samples generated from the Victim Model are directly applied to other models, testing transferability. Strongly transferable attack samples can hit almost all models in a black-box manner, which traditional white-box attacks~\cite{ebrahimi2018hotflip} can not match. Evaluation datasets like adversarial tasks of Adv-Glue~\cite{wang2021adversarial} showcase this transferability, aiding in robustness assessment. Relevant research~\cite{liu2016delving} endeavors to enhance the capability of adversarial example transferability by attacking ensemble models. It has been demonstrated~\cite{zheng2020efficient} that adversarial examples with better transferability can more effectively enhance the robustness of models in adversarial training. However, in the field of textual adversarial attacks, there has been a lack of in-depth research dedicated to how to improve the transferability of adversarial examples.

\subsection{Synchronization Work}
Prompt-Attack~\citep{xu2023llm} leverages the exceptional comprehension of LLMs and diverges from traditional adversarial methods. It employs a manual approach of constructing rule-based prompt inputs, requiring LLMs to output adversarial attack samples that can deceive itself and meet the modification rule conditions. This attack method achieved fully automatic and efficient generation of attack samples using the local model. However, the drawback is that the model may not perform the whole attacking process properly, resulting in mediocre attack effectiveness. Additionally, different prompts can significantly influence the quality of the model-generated attack samples. These generated attack samples, though somewhat transferable, fail to consider the model's internal reasoning, resulting in excessively high modification rates.
Our work shares similarities in that we both leverage the understanding capabilities of LLMs. However, we solely employ ChatGPT's language abstraction for Importance Level suggestions, maintaining the attack process within traditional text adversarial methods. Furthermore, our work is more focused on enhancing the transferability of attack samples and speeding up the attack, making them more widely applicable. This differs from the motivation of Prompt-Attack, which aims at the automated generation of samples that can deceive the model itself using LLMs.

\section{Method}

\subsection{Preliminaries: Adversarial Attack}

The task of adversarial attack aims at generating perturbations on inputs that can mislead the output of model. These perturbations can be very small, and imperceptible to human senses.

As for NLP tasks, given a corpus of $\textit{N}$ input texts, $\mathcal{X} = \{ x_1, x_2, x_3, \ldots , x_N \}$, and an output space $\mathcal{Y} = \{y_1, y_2, y_3, \ldots , y_N \}$ containing $\textit{K}$ labels, the language model F($\cdot$) learns a mapping $f:x \to y$ , which learns to classify each input sample $x \in X$ to the ground-truth label $y_{\text{gold}} \in \mathcal{Y}$:

\begin{equation}
\label{eq1}
F(x) = \arg\max\limits_{y_i \in \mathcal{Y}} P(y_i|x).
\end{equation}

The adversary of text $x \in X$ can be formulated as $x_{\text{adv}}$ = $x$ + $\epsilon$, where $\epsilon$ is a  perturbation to the input $x$. The goal is to mislead the victim language model F($\cdot$) within a certain constraint $C(x_{\text{adv}})$:

\begin{equation}
\label{eq2}
\begin{aligned}
    & F(x_{\text{adv}}) = \arg\max_{y_i \in \mathcal{Y}} P(y_i|x_{\text{adv}}) \neq F(x), \\
    &\text{and } \, C(x_{\text{adv}}, x), \leq \lambda
\end{aligned}
\end{equation}
where $\lambda$ is the coefficient, and C($x_{\text{adv}}$, x) is usually calculated by the semantic or syntactic similarity \citep{cer2018universal,oliva2011symss} between the input $x$ and its corresponding adversary $x_{\text{adv}}$.

\begin{figure}[t]
    \centering
    \includegraphics[width=\linewidth]{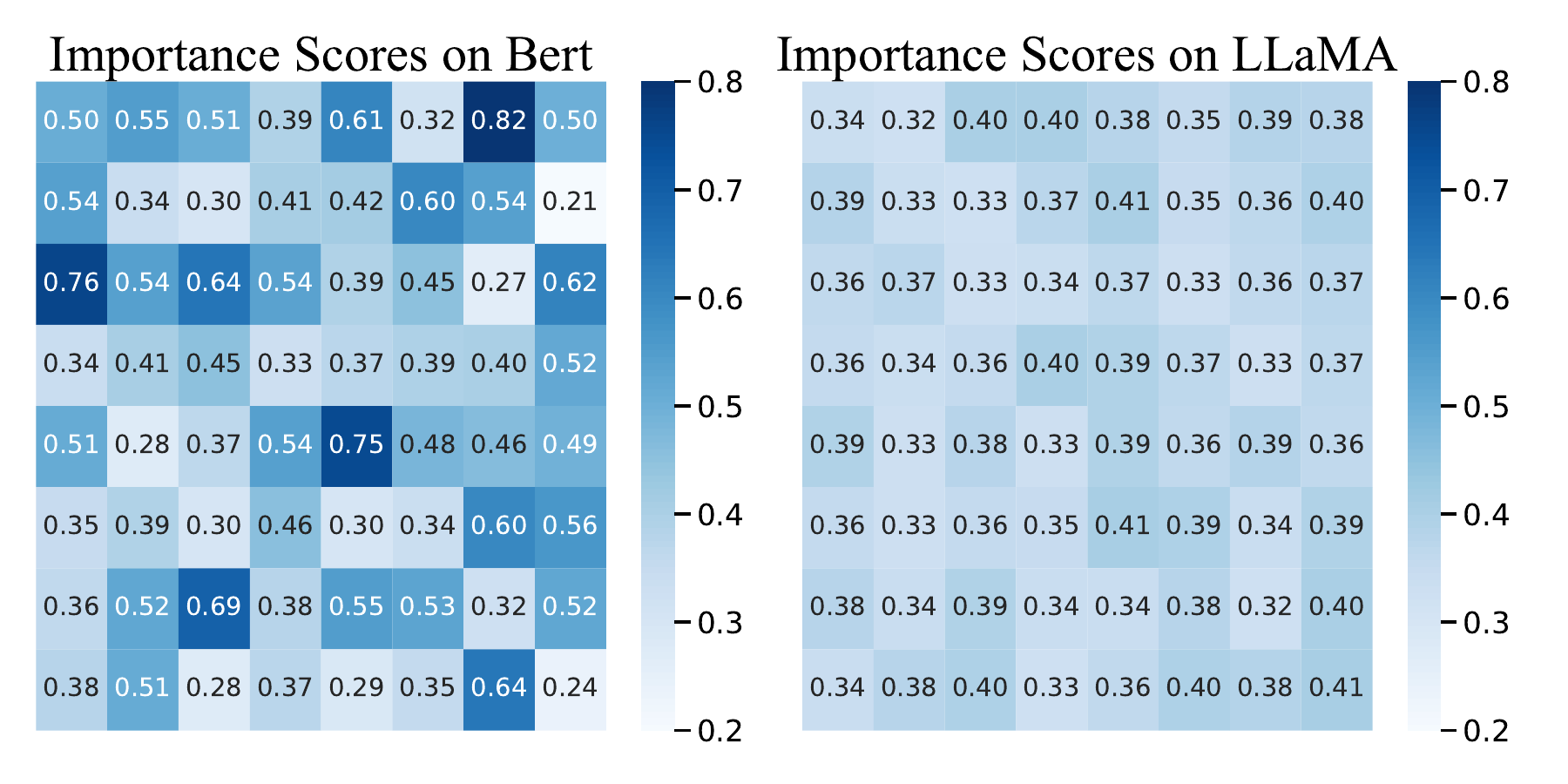}
    \caption{Importance score distribution of the same sentence given by BERT-Attack on BERT and LLaMA.}
    \label{fig:importscores}
\end{figure}

\begin{figure}[t]
    \centering
    \includegraphics[width=\linewidth]{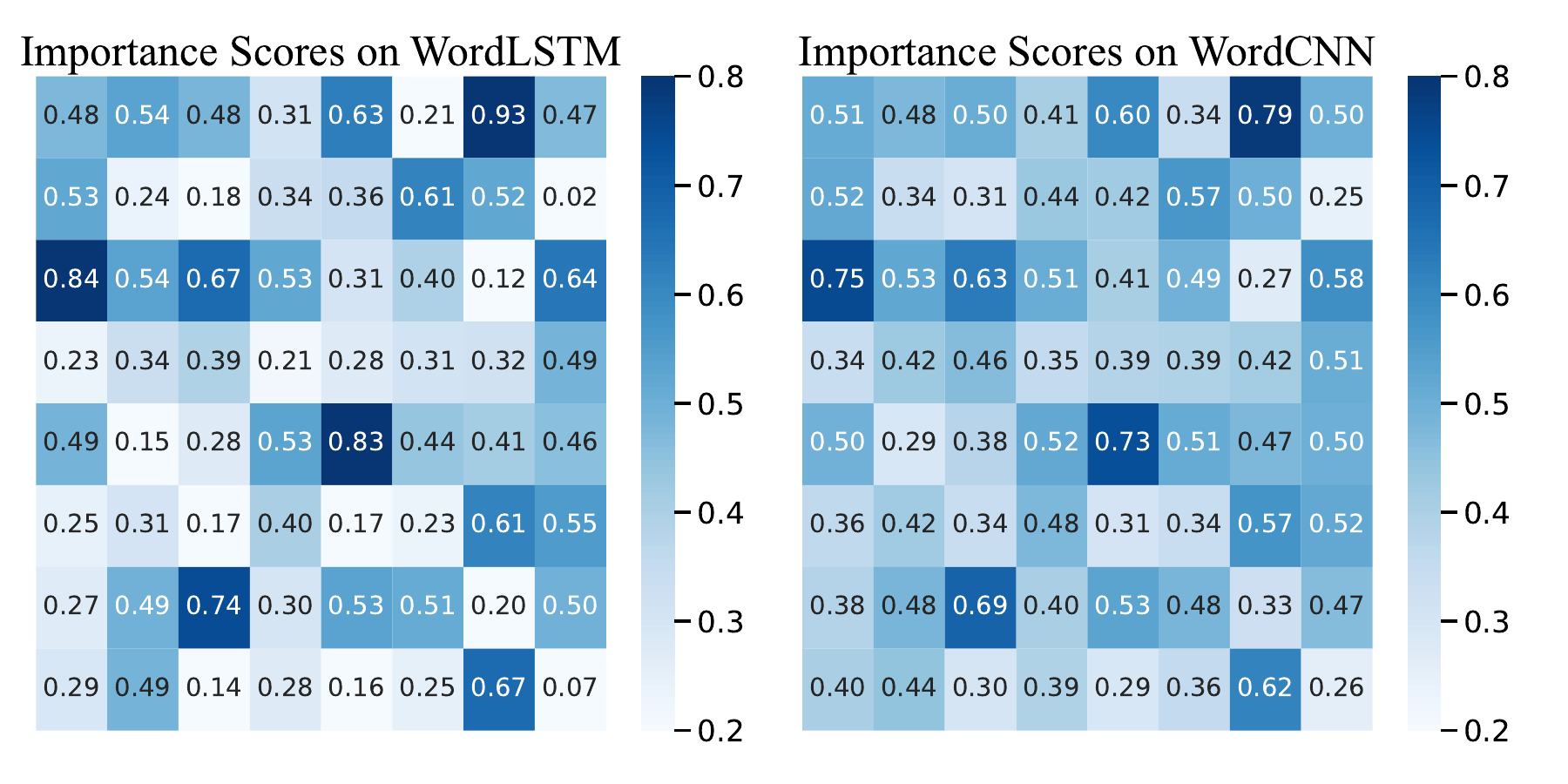}
    \caption{Importance score distribution of the same sentence given by BERT-Attack on WordCNN and WordLSTM.}
    \label{fig:other_is}
\end{figure}

\subsection{Limitations of Importance Score}
Existing predominant adversarial attack systems~\cite{morris2020textattack, jin2020bert, li2020bert} on LLMs typically adhere to a two-step process following BERT-Attack~\cite{li2020bert}. 
We thus take BERT-Attack as a representative method for analysis.
The core idea of BERT-Attack is to perform substitution according to \textbf{Importance Score}.
BERT-Attack calculates the importance score by individually masking each word in the input text, performing a single inference for each, and using changes in the confidence score and label of Victim Models to determine the impact of each word.
In BERT-Attack, importance score determine the subsequent attack sequence, which is crucial for the success of subsequent attacks and the times of attacks.

To explore the limited transferability of methods relying on importance scores, we analyze the importance score distribution among various models.
As illustrated in Figure \ref{fig:importscores}, there are substantial differences in the importance score derived from the same sentence when calculated using BERT and LLaMA. 
The former has a sharper distribution, while the latter essentially lacks substantial numerical differences. 
This phenomenon is consistent across multiple sentences.
Figure~\ref{fig:other_is} shows the different Importance Score distribution of the same sentence given by BERT-Attack on WordCNN and WordLSTM.

Given that importance score are crucial in the attack process of the aforementioned methods, variations in these scores can lead to entirely different adversarial samples. This variation explains the poor transferability of the generated attack samples, as demonstrated in Table~\ref{tab:bad_of_BA}.

\begin{figure}[t]
    \centering
    \includegraphics[width=\linewidth]{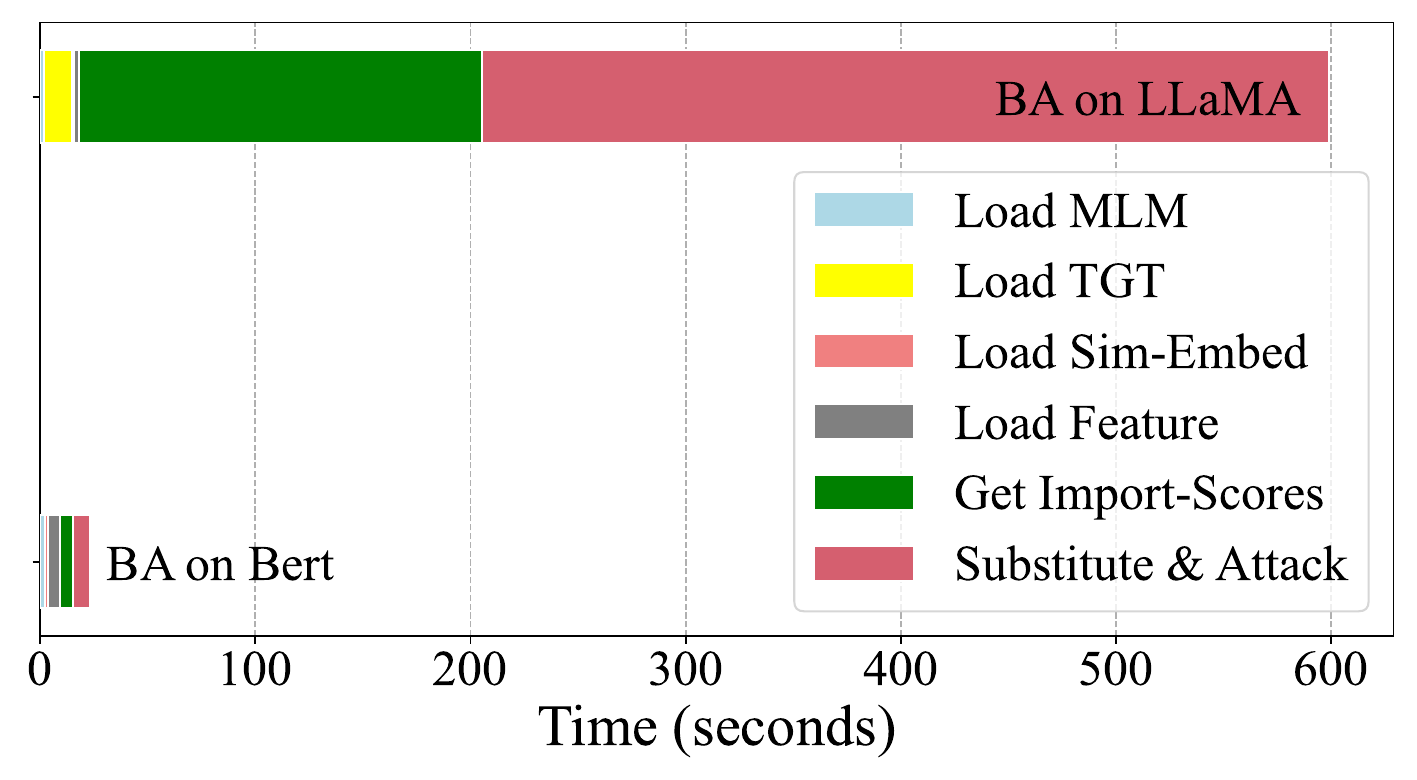}
    \caption{Time cost of each module from BERT-Attack on SA-LLaMA and BERT.}
    \label{fig:timecost}
\end{figure}

To investigate the causes of significant time overhead, we analyze the time consumption of various components involved when implementing a representative attacking method on both a small and a large model.
As depicted in Figure \ref{fig:timecost}, when applying BERT-Attack to attack small models like BERT, the average time spent per entry is very short, and the time consumption of various components is similar.
However, the time cost per inference on LLaMA far exceeds that of BERT, disrupting this balance. 
It is evident that in LLMs, over 80$\%$ of the time spent per entry in the attack is consumed by the  \textit{Get import-scores} and \textit{substitute \& Attack} component. 
In addition, it is worth noting that this phenomenon is even more pronounced in successfully attacked samples.
The underlying reason is that BERT-Attack necessitates performing the attack sequentially, according to the calculated importance score. 
This drawback becomes more pronounced when applied to large models, as both components require performing model inference, significantly increasing the time cost.

\begin{figure*}[t!]
    \centering
\includegraphics[width=\linewidth]{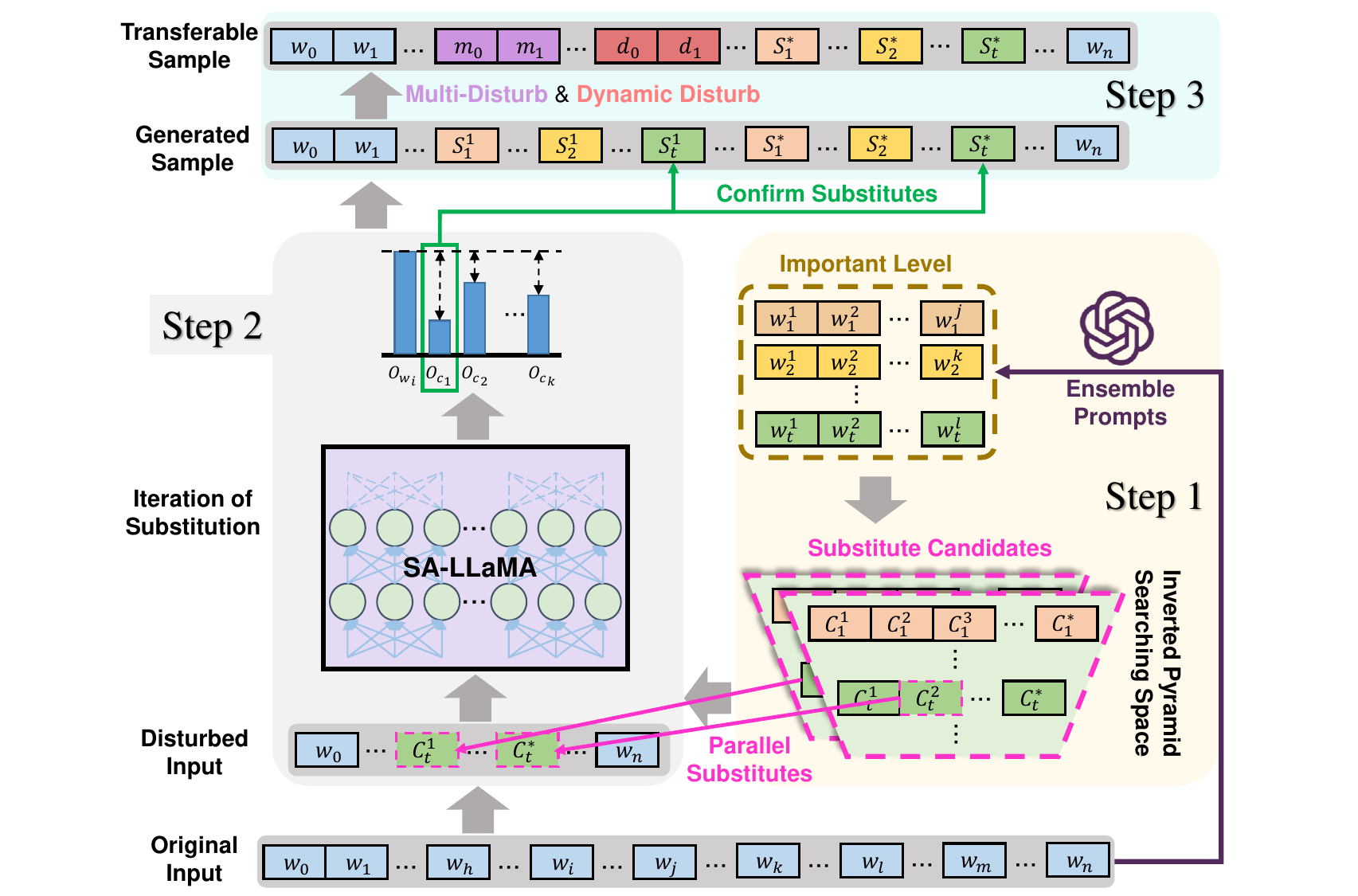}
    \caption{\textbf{Step 1:} Using ChatGPT to categorize words into 5 \textit{Important Level} with varying word counts. The \textit{Inverted Pyramid Searching Space} reflects the decreasing length of \textit{Substitute Candidates} based on decreasing levels. \textbf{Step 2:} Selecting words from the same level and generates a \textit{Disturbed Input} through \textit{Parallel Substitutions}. Exploring possible \textit{Disturbed Inputs} via \textit{SA-LLaMA}, choose the result surpassing the threshold as \textit{Generated Sample} from \textit{Confirmed Substitutions}. \textit{Substitution Iterations} will end when meet the finished condition. \textbf{Step 3:} Implementing \textit{Multi-Disturb} and \textit{Dynamic Disturb} produces \textit{Transferable Samples}.}
    \label{fig:framework}
\end{figure*}

\subsection{\method: Transferable and Fast Adversarial Attacks on LLMs}

To address the above limitations caused by the importance score mechanism, we introduce \method as a solution for transferable and fast adversarial attacks.
Firstly, as shown in Figure~\ref{fig:framework}, \method employs an external model as a third-party overseer to identify important units. 
In this way, \method avoids excessive dependence on the victim model during the attack process. 
Specifically, we design several instructions in Table~\ref{tab:prompt} for ChatGPT to partition all words in the original input according to their semantic importance. 
Another advantage of \method is its ability to utilize the rich semantic knowledge within ChatGPT, making the subsequent generation of attack sequences more universally semantic. 
In terms of speed, this approach does not require inference from the victim model, thus alleviating the problem of high inference costs for LLMs. 
\method only needs one inference to obtain a comprehensive Important Level priority queue, while the traditional approach requires inference times proportional to the length of the text.

Secondly, \method introduces the concept of \textbf{Importance Level} to facilitate parallel substitution.
Specifically, \method takes the original sentence as input and outputs a priority queue with 5 levels, with a different number of words in each level. 
The concept of importance level assumes that words within the same level have no specific order, enabling parallel replacement of candidate words at the same level. 
This parallel replacement process markedly decreases both the search space and the number of required inferences, offering a substantial improvement over the previous approach that relied on greedy, sequential word replacements.

\begin{table*}
\centering
\resizebox{\textwidth}{!}{
\begin{tabular}{c|c|l}
\hline
\begin{tabular}[c]{@{}c@{}}Level \end{tabular} & Abbre. & \multicolumn{1}{c}{ Perturbation Details} \\ \hline
\multirow{3}{*}{Character} & C1 & \begin{tabular}[c]{@{}l@{}}Choose at most two words in the sentence, and add letter to have typos.\end{tabular} \\ 
\cline{2-3} 
 & C2 & Change at most two letters in the sentence. \\ 
 \cline{2-3} 
 & C3 & Add at most two extraneous punctuation marks to the end of the sentence. \\ 
 \hline
\multirow{3}{*}{Word} & W1 & Replace at most two words in the sentence with synonyms. \\ 
\cline{2-3} 
 & W2 & \begin{tabular}[c]{@{}l@{}}Delete at most two words in the sentence with synonyms.\end{tabular} \\ \cline{2-3} 
 & W3 & Add at most two semantically neutral words to the sentence. \\ \hline
\multirow{3}{*}{Sentence} & S1 & \begin{tabular}[c]{@{}l@{}}Add a randomly generated short meaningless handle like @fasuv3. \end{tabular} \\ \cline{2-3} 
 & S2 & Change the syntactic structure and word order of the sentence. \\ \cline{2-3} 
 & S3 & Paraphrase the sentence with ChatGPT. \\ \hline
\end{tabular}
}
\caption{Three types of nine different disturbance methods.}
\label{tab:AG}
\end{table*}

Additionally, we employ a \textit{reverse pyramid search space} strategy for importance levels, optimizing the search space to reduce inefficient search expenditures. Words prioritized at higher levels are presumed to significantly influence sentence sentiment. Consequently, a larger search space is utilized to identify semantically similar words, with the aim of replacing them with synonyms that precipitate a notable decline in the performance of victim model. For words at lower priority levels, a smaller search space suffices, as alterations to these words minimally impact sentence sentiment. Excessive searching at these levels can lead to increased inference costs without substantially enhancing effectiveness.

\subsection{Multi-Disturb \& Dynamic-Disturb}

Building on the aforementioned method, to further enhance the robustness of adversarial attack samples, we strategically propose two tricks for optimization. Following \cite{xu2023llm}, we use 9 ways of disturbance, including character-level, word-level, and sentence-level disturbances in Table~\ref{tab:AG}. However, how to set the ratios of these three types of disturbances largely determines the quality of the transferability from generated attack samples. Therefore, the following strategy is proposed.

In the step of evaluating whether an attack sample is effective, traditional attack methods almost solely rely on the confidence of model output, a practice that undoubtedly promotes overfitting of attack samples to the model architecture. Therefore, in the process of determining the effectiveness of a replacement, we introduce random disturbance to the decrease in model confidence. This may result in the loss of some already successfully attacked samples, but it also prevents the occurrence of the phenomenon where the attack stops after succeeding on this particular Victim model. Traditional methods rely heavily on model confidence, leading to overfitting. To counter this, we introduce random disturbances during effectiveness assessment, reducing model confidence. This might sacrifice past successful attacks but prevents reliance on the success of the victim model.

These two strategies can be incorporated into traditional text adversarial attack methods as a post-processing step, significantly improving the transferability of adversarial samples. Specifically, the Multi-Disturb strategy refers to introducing a variety of disturbances within the same sentence. Table~\ref{tab:AG} outlines 9 ways of disturbance, including character-level, word-level, and sentence-level disturbances, which can greatly enhance the transferability of attack samples.
Dynamic-Disturb refers to using an FFN+Softmax network to assess the length and structural distribution of the input sentence, outputting the ratios of these three types of disturbances.

In assessing attack sample effectiveness, traditional methods heavily depend on model output confidence, likely leading to overfitting to the model architecture. To rectify the problem, we incorporate random disturbance to diminish model confidence during replacement evaluation. Our experiments confirm that this two tricks can be adapted to almost all text adversarial attack methods, significantly enhancing transferability and increasing the ability of adversarial samples to confuse models through adaptive post-processing.

\begin{table}
\small
\centering
\begin{tabular}{ccccc}
\toprule
\textbf{Dataset} & \textbf{Train} & \textbf{Test} & \textbf{Avg Len} & \textbf{Classes} \\
\midrule
Yelp &  560k & 38k & 152&2  \\
IMDB & 25k & 25k & 215&2  \\
AG's News& 120k & 7.6k & 73&4\\
MR & 9k & 1k & 20&2\\
SST-2 & 7k & 1k & 17 & 2 \\
\bottomrule
\end{tabular}
\caption{Overall statistics of datasets.}
\label{tab:overall_datasets}
\end{table}

\begin{table*}[t!]
\centering
\small
\resizebox{\linewidth}{!}{
\begin{tabular}{llrrr|llrrr}
\toprule
\textbf{Dataset} & \textbf{Method} &\textbf{A-rate}$\uparrow$ & \textbf{Mod}$\downarrow$ & \textbf{Sim}$\uparrow$ &\textbf{Dataset} & \textbf{Method} & \textbf{A-rate }$\uparrow$& \textbf{Mod} $\downarrow$& \textbf{Sim}$\uparrow$\\
\midrule

\multirow{7}{*}{\textbf{Yelp}} & TextFooler & 78.9 & \underline{9.1} & \underline{0.73}  & \multirow{7}{*}{\textbf{IMDB}} & TextFooler & 83.3 & \underline{8.1} & 0.79 \\
& BERT-Attack        & 80.5 & 11.5 & 0.69 & & BERT-Attack        & 84.2 & 9.6  & 0.78 \\
& SDM-Attack        & 81.1 & 10.7 & 0.71 & & SDM-Attack        & 86.1 & 8.9  & 0.75 \\
& \method(Zero-Shot) & 81.3 & 10.3 & 0.71 & & \method(Zero-Shot) & 86.1 & 8.7  & \underline{0.81} \\
& \method(Few-Shot)  & 83.7 & \textbf{9.0}  & \textbf{0.77} & & \method(Few-Shot)  & 86.7 & \textbf{7.4}  & 0.76 \\
& +MD                & \textbf{84.6} & 12.3 & 0.71 & & +MD                & \underline{87.1} & 10.7 & 0.77 \\
& +MD\hspace{1em}+DD & \underline{84.5} & 11.9 & 0.72 & & +MD\hspace{1em}+DD & \textbf{87.7} & 9.9  & \textbf{0.81} \\
\midrule
\multirow{7}{*}{\textbf{AG's News}} & TextFooler & 73.2 & \underline{16.1} & 0.54 & \multirow{7}{*}{\textbf{MR}} & TextFooler& 81.3 & 10.5 & 0.53  \\
& BERT-Attack        & 76.6 & 17.3 & \underline{0.59} & & BERT-Attack        & 82.8 & 10.2 & 0.51 \\
& SDM-Attack        & 79.3 & 16.2 & \textbf{0.61} & & SDM-Attack        & 84.0 & \underline{9.9} & \underline{0.55} \\
& \method(Zero-Shot) & 77.1 & 18.3 & 0.52 & & \method(Zero-Shot) & \textbf{84.6} & \textbf{8.9} & \textbf{0.58} \\
& \method(Few-Shot)  & 81.9 & \textbf{16.1} & 0.58 & & \method(Few-Shot)  & 83.1 & 12.4 & 0.44 \\
& +MD                & \underline{82.8} & 19.4 & 0.53 & & +MD                & 83.0 & 13.2 & 0.40 \\
& +MD\hspace{1em}+DD & \textbf{83.0} & 19.1 & 0.55 & & +MD\hspace{1em}+DD & \underline{84.0} & 10.4 & 0.50 \\
\midrule
\multirow{7}{*}{\textbf{SNLI}} & TextFooler & 82.8 & 14.1 & 0.39 & \multirow{7}{*}{\textbf{MNLI}} & TextFooler & 77.1 & 11.5 & 0.51 \\
& BERT-Attack        & 80.5 & 12.3 & 0.46 & & BERT-Attack    & 75.8 & 8.4 & 0.54 \\
& SDM-Attack        & 83.1 & 14.6 & 0.42 & & SDM-Attack & 79.3 & 10.2 & 0.55 \\
& \method(Zero-Shot) & 82.9	& \textbf{10.2} & 0.45 & & \method(Zero-Shot) & \underline{79.3} & \underline{8.4} & \underline{0.55} \\
& \method(Few-Shot)  & 82.7 & \underline{10.4} & \textbf{0.47} & & \method(Few-Shot)  & 78.2 & 7.9 & \textbf{0.57} \\
& +MD                & \textbf{83.6} & 11.7 & \underline{0.46} & & +MD                & 77.7 & \textbf{8.3} & 0.53  \\
& +MD\hspace{1em}+DD & \underline{83.5} & 10.9 & 0.41 & & +MD\hspace{1em}+DD & \textbf{79.4} & 8.5 & 0.52 \\
\bottomrule
\end{tabular}}
\caption{Automatic evaluation results of attack success rate (A-rate), modification rate (Mod), and semantic similarity (Sim) on SA-LLaMA. $\uparrow$ represents the higher the better and $\downarrow$ means the opposite. The best results are \textbf{bolded}, and the second-best
ones are \underline{underlined}.}
\label{tab:main}
\end{table*}

\section{Experiments}

\subsection{Experimental Setups}



\paragraph{Tasks and Datasets} Following \cite{li2020bert}, we evaluate the effectiveness of the proposed \method on classification tasks upon diverse datasets covering news topics (AG's News;~\citealp{zhang2015character}), sentiment analysis at sentence (MR;~\citealp{pang2005seeing}) and document levels (IMDB\footnote{https://datasets.imdbws.com/} and Yelp Polarity;~\citealp{zhang2015character}).  
As for textual entailment, we use a dataset of sentence pairs (SNLI;~\citealp{bowman2015large}) and a dataset with multi-genre (MultiNLI;~\citealp{williams2018broad}). Following~\citep{jin2020bert, alzantot2018generating}, we attack 1k samples randomly selected from the test set of each task.
The statistics of datasets and more details can be found in Table~\ref{tab:overall_datasets}. 

\paragraph{Baselines} We compare \method with recent studies: 1) TextFooler~\citep{jin2020bert}, which finds important words via probability-weighted word saliency and then applies substitution with counter-fitted word embeddings.  2) BERT-Attack~\citep{li2020bert}, which uses a mask-predict approach to generate adversaries. 3) SDM-Attack~\citep{fang2023modeling}, which employs reinforcement learning to determine the attack sequence.
We use the official codes BERT-Attack and TextAttack tools \cite{morris2020textattack} to perform attacks in our experiments. 
The \method(zero-shot) denotes that no demonstration examples were provided when generating the \textit{Important Level} with ChatGPT. 
Conversely, the \method(few-shot) uses five demonstrations as context information. 
To ensure a fair comparison, we follow \cite{morris2020textattack} to apply constraints for \method. 

\paragraph{Implementation Details} Following established training protocols, we fine-tuned a LLaMA-2-7B model to develop specialized Task-LLaMA models tailored for specific downstream tasks.
Among them, the Task-LLaMA fine-tuned on the IMDB training set achieved an accuracy of 96.95\% on the test set, surpassing XLNET with additional data, which achieved 96.21\%. The model achieved 93.63\% on another sentiment classification dataset, SST-2, indicating that Task-LLaMA is not overfitting to the training data but a strong baseline for experiments.

\paragraph{Automatic Evaluation Metrics} Following prior work \citep{jin2020bert, morris2020textattack}, we assess the results with the following metrics: 1) attack success rate (A-rate): post-attack model performance decline; 2) Modification rate (Mod): percentage of altered words compared to the original; 3) Semantic similarity (Sim): cosine similarity between original and adversary texts via universal sentence encoder (USE; \citealp{cer2018universal}); and 4) Transferability (Trans): the mean accuracy decreases across three models between adversarial and original samples.

\paragraph{Manual Evaluation Metrics} Following \citep{fang2023modeling}, We further manually validate the quality of the adversaries from three challenging properties. 1) Human prediction consistency (Con): how often human judgment aligns with the true label; 2) Language fluency (Flu): measured on a scale of 1 to 5 for sentence coherence \cite{gagnon2019salsa}; and 3) Semantic similarity ($\text{Sim}_\text{hum}$): gauging consistency between input-adversary pairs, with 1 indicating \textit{agreement}, 0.5 \textit{ambiguity}, and 0 \textit{inconsistency}.

\subsection{Overall Performance}
Table~\ref{tab:main} shows the performance of different systems on four benchmarks.
As shown in Table \ref{tab:main}, \method consistently achieves the highest attack success rate to attack LLaMA and has little negative impact on Mod and Sim.
Additionally, \method mostly obtains the best performance of modification and similarity metrics, except for AG's News, where \method achieves the second-best.
In general, our method can simultaneously satisfy the high attack success rate with a lower modification rate and higher similarity.
We additionally observe that \method achieves a better attack effect on the binary classification task. Empirically, when there exist more than two categories, the impact of each replacement word may be biased towards a different class, leading to an increase in the perturbation rate.

\label{app:random_attack}
In Table~\ref{tab:random}, we evaluate the effectiveness of attack order. Utilizing a random attack sequence leads to a reduction in the success rate of attacks and a significant increase in the modification rate, as well as severe disruption of sentence similarity. This implies that each attack path is random, and a substantial amount of inference overhead is wasted on futile attempts. Although we adhere to the threshold constraints of the traditional adversarial text attack domain, the text can still be successfully attacked. However, under conditions of high modification rates and low similarity, the text has been altered significantly from its original semantics, contravening the purpose of the task. Furthermore, a random attack sequence incurs a substantial additional cost in terms of attack speed, resulting in a nearly doubled time delay.

\begin{table}[t!]
    \small
    \centering
    \begin{tabular}{lcccc}
    \toprule
         \textbf{Method} & \textbf{A-rate} & \textbf{Mod}$\downarrow$  & \textbf{Sim}$\uparrow$ & 
         \textbf{Time Cost} $\downarrow$\\
    
    \midrule
    BA-IS & 84.2 & 9.9 & 0.78 & 11.1 \\
    TF-IS & 86.3 & 10.4 & 0.77 & 10.4 \\
    BA-IL & \underline{86.9} & \textbf{9.2} & \textbf{0.83} & \underline{3.3} \\
    TF-IL & \textbf{87.7} & \underline{9.9} & \underline{0.81} & \textbf{2.7} \\
    BA-RD & 75.3 & 18.4 & 0.47 & 27.6 \\
    TF-RD & 77.8 & 17.4 & 0.51 & 25.9 \\

    \bottomrule
    \end{tabular}
    \caption{Automatic evaluation results of attack success rate (A-rate), modification rate (Mod), semantic similarity (Sim) and time cost (Time Cost) on SA-LLaMA. $\uparrow$ represents the higher the better and $\downarrow$ means the opposite. The best results are \textbf{bolded}, and the second-best ones are \underline{underlined}. BA means BERT-Attack and TF means our method. IS means Important Score, IF means Important Level and RD means random attacking.}
    \label{tab:random}
\end{table}

\begin{table}[t!]
    \centering
    \small
    \scalebox{.9}{
    \begin{tabular}{l@{\hspace{2pt}}cccc@{\hspace{2.2pt}}c@{\hspace{2.2pt}}}
    \toprule
         & \textbf{CNN} & \textbf{LSTM} & \textbf{BERT} & \textbf{LLaMA} & \textbf{Baichuan}\\
    \midrule
    \textbf{CNN}     & \ \ -2.2 & +33.8 & +58.5 & +37.4 & +24.0 \\
    \textbf{LSTM}    & +22.4 & \ \ -3.8 & +31.2 & +45.8 & +29.4 \\
    \textbf{BERT}     & +21.4 & +26.0 & \ \ +2.2 & +25.4 & +25.6 \\
    \textbf{LLaMA}    & +24.0 & +26.8 & +41.8 & +12.4 & +24.4 \\
    \textbf{Baichuan} & +22.2 & +20.6 & +22.4 & +21.6 & \ \ +4.4 \\ 
    \textbf{ChatGPT}  & +14.8 & +18.6 & +22.4 & +22.4 & +18.8 \\ 
    \midrule
    \textbf{Average}  & +21.2 & +23.6 & +35.2 & +30.0 & +27.6\\ 
    \bottomrule
    \end{tabular}
    }
    \caption{Transferability evaluation of \method Samples on IMDB dataset. Each element is calculated from the difference in Attack Success Rate between \method and BERT-Attack.}
    \label{tab:trans_of_FA}
\end{table}

\subsection{Transferability}
We evaluate the transferability of \method samples to detect whether the samples generated from \method can effectively attack other models. We conduct experiments on the IMDB datasets and use BERT-Attack as a baseline.
Table~\ref{tab:trans_of_FA} shows the improvement of the attack success rate of \method over BERT-Attack.
It can be observed that \method achieves obvious improvements over BERT-Attack when applied to other models.
In particular, the new samples generated by \method can lower the accuracy by over 10 percent on binary classification tasks, essentially confusing the victim model. 
Even a powerful baseline like ChatGPT would drop to only 68.6\% accuracy. 
It is important to highlight that these samples do not necessitate attacks tailored to specific victim models.

\subsection{Efficiency}
\begin{figure}[t!]
\includegraphics[width=\linewidth]{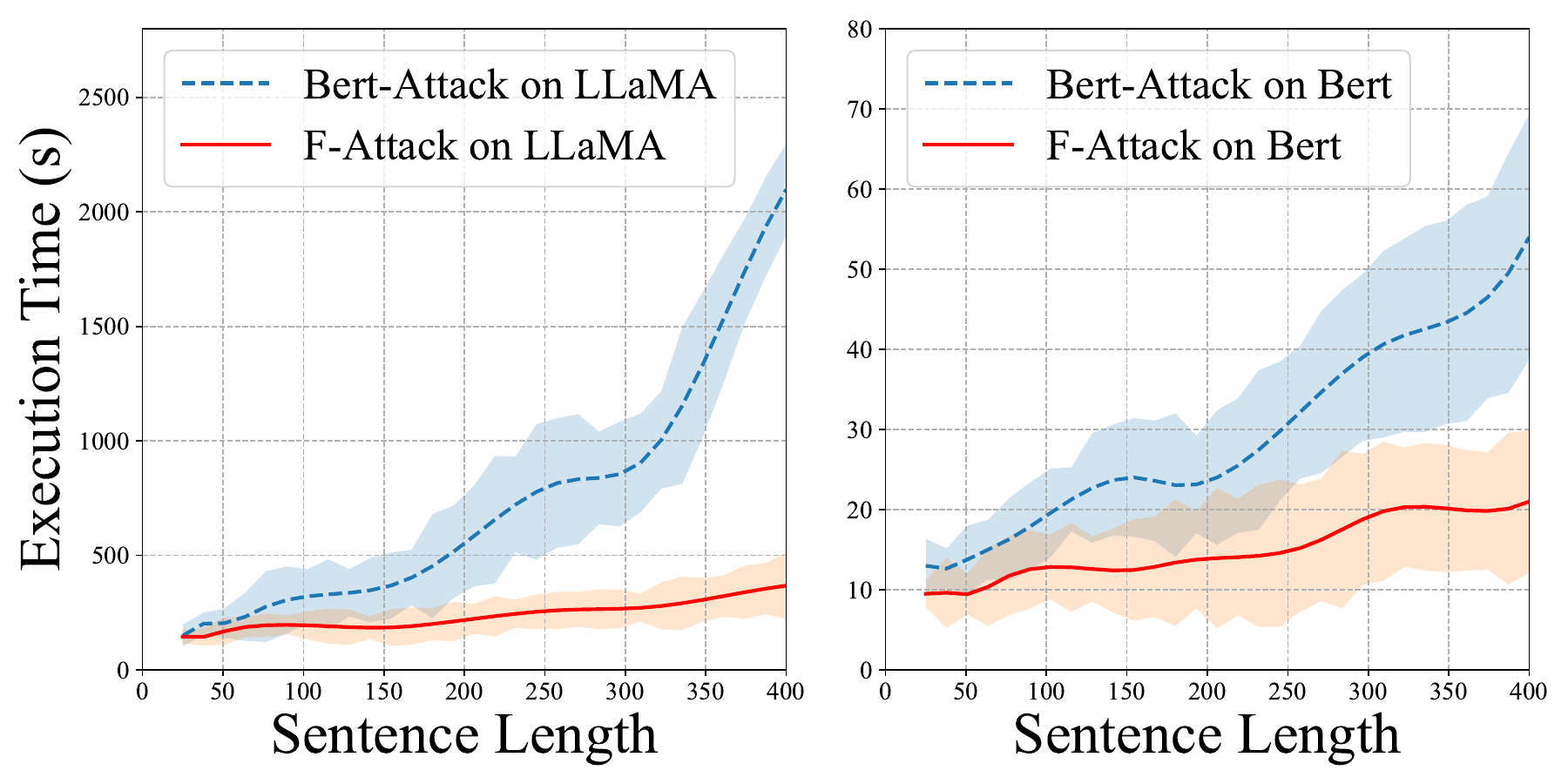}
\caption{The time cost according to varying sentence lengths in the IMDB dataset. The left is on LLaMA while the right is on BERT.} 
\label{fig:speed}
\end{figure}

We probe the efficiency according to varying sentence lengths in the IMDB dataset.
As shown in Figure~\ref{fig:speed}, the time cost of \method is surprisingly mostly better than BERT-Attack, which mainly targets obtaining cheaper computation costs with lower attack success rates in Table~\ref{tab:main}. 
Furthermore, with the increase of sentence lengths, \method maintains a stable time cost, while the time cost of BERT-Attack is exploding. 
The reason is that \method has the advantage of much faster parallel substitution, hence as the sentence grows, the increase in time cost will be much smaller. 
These phenomena verify the efficiency advantage of \method, especially in dealing with long texts.

BERT-Attack has emerged as the most widely adopted and efficacious technique for adversarial text attacks in the existing literature. Our \method has conducted a comprehensive analysis and introduced innovative enhancements to this approach. Consequently, we have chosen to benchmark our performance metrics against BERT-Attack to enable a more straightforward and direct comparative assessment. We have supplemented the speed experiments related to TextFooler and SDM-Attack on the IMDB dataset attacking SA-LLaMA, with the results pertaining to Figure~\ref{fig:speed} on Table~\ref{tab:a_speed}.

\begin{table}[t!]
    \small
    \centering
    \begin{tabular}{lcc}
    \toprule
    \textbf{Method} & \textbf{on SA-LLaMA} & \textbf{on Bert} \\
    \midrule
    TextFooler & 1034 & 32 \\
    Bert-Attack & 1126 & 29 \\
    SDM-Attack & 983 & 21 \\
    TF-Attack & 123 & 14 \\
    \bottomrule
    \end{tabular}
    \caption{The average attacking speed(s/item) comparison of different methods.}
    \label{tab:a_speed}
\end{table}

\begin{table}[t!]
    \small
    \centering
    \begin{tabular}{lrrrr}
    \toprule
    \textbf{Method} & \textbf{A-rate}$\uparrow$ & \textbf{Mod}$\downarrow$ & \textbf{Sim}$\uparrow$ \\
    \midrule
    TextBugger & 73.9 & 10.3 & 0.69 \\
    TextFooler & 78.9 & 9.1 & 0.73 \\
    Bert-Attack & 80.5 & 11.5 & 0.69 \\
    SDM-Attack & 81.1 & 10.7 & 0.71 \\
    Zero-Shot & 81.3 & 10.3 & 0.71 \\
    Few-Shot & 83.7 & 9.0 & 0.77 \\
    +MD & 84.6 & 12.3 & 0.71 \\
    +MD +DD & 84.5 & 11.9 & 0.72 \\
    \bottomrule
    \end{tabular}
    \caption{More baseline of adversarial attack methods results.}
    \label{tab:more_baseline}
\end{table}

\begin{table}[t!]
    \small
    \centering
    \begin{tabular}{llrrc}
    \toprule
         \textbf{Dataset} &  & \textbf{Con}$\uparrow$  & \textbf{Flu}$\uparrow$ & 
         $\textbf{Sim}_\text{hum}$ $\uparrow$\\
    
    \midrule
    \multirow{2}{*}{\textbf{IMDB}} & Original  & 0.88  & 4.0 & \multirow{2}{*}{0.82}\\
    & \method & 0.79 & 3.8 &  \\
    
     \midrule
    \multirow{2}{*}{\textbf{MR}} & Original  & 0.93  & 4.5 & \multirow{2}{*}{0.93}\\
    & \method & 0.87 & 4.1 &  \\
    
    \bottomrule
    \end{tabular}
    \caption{Human evaluation results comparing the original input and generated adversary by \method of human prediction consistency (Con), language fluency (Flu), and semantic similarity ($\text{Sim}_\text{hum}$).}
    \label{tab:manual}
\end{table}

\subsection{Manual evaluation}
We follow \citealp{fang2023modeling} to perform manual evaluation. At the beginning of manual evaluation, we provided some data to allow crowdsourcing workers to unify the evaluation standards. We also remove the data with large differences when calculating the average value to ensure the reliability and accuracy of the evaluation results.
We first randomly select 100 samples from successful adversaries in IMDB and MR datasets and then ask ten crowd-workers to evaluate the quality of the original inputs and our generated adversaries. The results are shown in Table \ref{tab:manual}. For human prediction consistency, humans can accurately judge 93\% of the original inputs on the IMDB dataset while maintaining an 87\% accuracy rate with our generated adversarial examples. This suggests that \method can effectively mislead LLMs without altering human judgment. Regarding language fluency, the scores of our adversarial examples are comparable to the original inputs, with a minor score difference of no more than 0.3 across both datasets. Moreover, the semantic similarity scores between the original inputs and our generated adversarial examples stand at 0.93 for IMDB and 0.82 for MR, respectively. The result on NLI task is in Table~\ref{tab:manual_all}. Overall, \method successfully preserves these three essential attributes.

\subsection{More baseline experiments}

BERT-Attack and TextFooler are currently the most widely used and powerful baselines in the field of text adversarial attacks, so we initially compared only these two methods. Due to time constraints, we conducted relevant experiments on the YELP dataset using TextBugger, with the victim model being SA-LLaMA. The results pertaining to Table~\ref{tab:more_baseline} are supplemented in Table~\ref{tab:more_baseline}:

\begin{table}[t!]
    \small
    \centering
    \scalebox{.9}{
    \begin{tabular}{llrrc}
    \toprule
         \textbf{Dataset} &  & \textbf{Con}$\uparrow$  & \textbf{Flu}$\uparrow$ & 
         $\textbf{Sim}_\text{hum}$ $\uparrow$\\
    
    \midrule
    \multirow{4}{*}{\textbf{MNLI}} & Original  & 0.96  & 4.6 & 1.00\\
    & TextFooler & 0.86  & 4.1 & 0.85\\
    & BERT-Attack & 0.82  & 4.3 & 0.91\\
    & \method & 0.91 & 4.6 & 0.91 \\
    
     \midrule
    \multirow{4}{*}{\textbf{SNLI}} & Original  & 0.85  & 4.6 & 1.00\\
     & TextFooler & 0.77  & 3.9 & 0.84\\
     & BERT-Attack & 0.78  & 3.7 & 0.84\\
    & \method & 0.76 & 4.3 & 0.81 \\
    
    \bottomrule
    \end{tabular}
    }
    \caption{Human evaluation results regarding human prediction consistency (Con), language fluency (Flu), and semantic similarity ($\text{Sim}_\text{hum}$).}
    \label{tab:manual_all}
\end{table}

\begin{table}[t!]
    \centering
    \small
    \begin{tabular}{l@{\hspace{1.6pt}}cccc@{\hspace{1.4pt}}c@{\hspace{2pt}}}
    \toprule
    
    ASR$\uparrow$ & LLaMA-2b & LLaMA-2c & ChatGPT & Claude \\
    \midrule
    Prompt1 & / & / & 86.54\% & 85.46\% \\
    Prompt2 & / & 12.28\% & 88.68\% & 78.64\% \\
    Prompt3 & / & / & 79.74\% & 71.07\% \\
    Prompt4 & / & 11.22\% & 84.36\% & 73.76\% \\
    Prompt5 & 13.67\% & / & 83.16\% & 77.81\% \\
    \bottomrule
    \end{tabular}
    \caption{The results of \method by different Prompts on different models.}
    \label{tab:prompt}
\end{table}

\begin{table*}
\centering
\begin{tabular}{p{\linewidth} }\textbf{Prompt}\\
\hline

Rank each word in the input sentence into five levels based on its determining influence on the overall sentiment of the sentence.  \\\hline
Determine the impact of each word on the overall sentiment of the sentence and categorise it into 5 levels. \\\hline
Rank the words from most to least influential in terms of their impact on the emotional tone of the sentence at 5 levels. \\\hline
Please classify each word into five levels, based on their importance to the overall emotional classification of the utterance. \\\hline
Assign each word to one of five levels of importance based on its contribution to the overall sentiment: Very High, High, Moderate, Low, Very Low. \\\hline

\end{tabular}
\caption{Different prompts used on ChatGPT to generate Important Level
}
\label{tab:prompts}
\end{table*}


In fact, in earlier experiments Table \ref{tab:prompt}, we attempted to use local LLMs or other different architectures of LLMs as \textit{selectors for attack order} but found that open-source LLMs base models such as 7B, 13B, or even 30B could not understand our instructional intent well in this specific scenario, no matter Base models or Chat models. If we were to introduce a specially designed task fine-tuning process to achieve this functionality, it would not only require a large amount of manually labeled dataset but also incur even larger model training costs, which contradicts one of our motivations, accelerating adversarial attacks on large models. And we have supplemented the experimental results using Claude as a third-party selector, using the same Ensemble Prompt as ChatGPT. The experimental results have been added to Appendix. The following Table shows 100 samples attacked through different Prompts on different LLMs, '/' represents that LLM can not correctly output the Important Level. Table \ref{tab:prompts} shows all the 5 prompts we use in experiments.

\begin{table}[t!]
    \centering
    \small
    \scalebox{.9}{
    \begin{tabular}{lrrrr}
    \toprule
    \textbf{Victim models} &\textbf{A-rate}$\uparrow$ & \textbf{Mod}$\downarrow$ & \textbf{Sim}$\uparrow $& \textbf{Trans}$\downarrow$ \\
    \midrule
    \textbf{WordCNN}  & 96.3 & 9.1  & 0.84 & 78.5   \\
    \textbf{WordLSTM} & 92.8 & 9.3  & 0.85 & 75.1   \\
    \textbf{BERT}     & 90.6 & 9.9  & 0.81 & 70.2   \\
    \textbf{LLaMA}    & 91.8 & 13.1 & 0.74 & 68.6  \\
    \textbf{Baichuan} & 92.5 & 11.8 & 0.75 & 71.4 \\ 
    
    \bottomrule
    \end{tabular}}
    \caption{\method against other models.}
    \label{tab:other}
\end{table}

\section{Analysis}

\subsection{Generalization to More Victim Models}\label{sec:models}
Table~\ref{tab:other} shows that \method not only has better attack effects against WordCNN and WordLSTM, but also misleads BERT and Baichuan, which are more robust models.
For example, on the IMDB datasets, the attack success rate is up to 92.5\% against Baichuan with a modification rate of only about 11.8\% and a high semantic similarity of 0.75. Furthermore, the model generated by the Victim model created a decrease in accuracy to 71.4\% on various black-box models of different scales.

\subsection{Adversarial Training} 
Following \citealp{fang2023modeling}, we further investigate to improve the robustness of victim models via adversarial training using the generated adversarial samples. Specifically, we fine-tune the victim model with both original training datasets and our generated adversaries and evaluate it on the same test set.
As shown in Table \ref{tab:training}, compared to the results with the original training datasets, adversarial training with our generated adversaries can maintain close accuracy, while improving performance on attack success rates, modification rates, and semantic similarity. The victim models with adversarial training are more difficult to attack, which indicates that our generated adversaries have the potential to serve as supplementary corpora to enhance the robustness of victim models.

\begin{table}[t!]
    \small
    \centering
    \begin{tabular}{lrrrr}
    \toprule
    \textbf{Dataset} & \textbf{Acc}$\uparrow$  & \textbf{A-rate}$\uparrow$ & \textbf{Mod}$\downarrow$ & \textbf{Sim}$\uparrow$ \\
      \midrule
     \textbf{Yelp} & 97.4 & 81.3 & 8.5 & 0.73  \\
     +Adv Train    & 95.9 & 65.7 & 12.3 & 0.67 \\ 
     \midrule
     \textbf{IMDB} & 97.2 & 86.1 & 4.6 & 0.81 \\
     +Adv Train    & 95.5 & 70.2 & 7.3 & 0.78 \\
    \midrule
    \textbf{AG-NEWS}  & 95.3 & 77.1 & 15.3 & 0.83 \\
    +Adv Train        & 85.1 & 75.3 & 23.3 & 0.61 \\
    \midrule
    \textbf{MR} & 95.9 & 83.2 & 11.1 & 0.53\\
     +Adv Train & 91.7 & 71.8 & 14.6 & 0.67 \\
     \midrule
     \textbf{SST-2}  & 97.1 & 89.7 & 14.3 & 0.85 \\
      +Adv Train     & 92.2 & 68.6 & 16.8 & 0.83 \\
     \bottomrule
   \end{tabular}
    \caption{Adversarial training results.}
    \label{tab:training}
\end{table}

\begin{table*}[t!]
\centering
\small
\begin{tabular}{lp{8cm}cccc}
\toprule 
\textbf{Method}  & \textbf{Text} (MR; Negative)& \textbf{Result} & \textbf{Mod}$\downarrow$ & \textbf{Sim}$\uparrow$ & \textbf{Flu}$\uparrow$ \\
\midrule
\multirow{2}{*}{Original} & Davis is so enamored of her own creation that she can not see how insufferable the character is. & \multirow{2}{*}{-}& \multirow{2}{*}{-}  & \multirow{2}{*}{-} & \multirow{2}{*}{5} \\
\midrule
 \multirow{2}{*}{TextFooler} & Davis is \textcolor{blue}{well} enamored of her own \textcolor{blue}{infancy} that she \textcolor{blue}{could}  not \textcolor{blue}{admire} how \textcolor{blue}{infernal} the \textcolor{blue}{idiosyncrasies} is. &\multirow{2}{*}{\textit{Success}} &\multirow{2}{*}{33.3} & \multirow{2}{*}{0.23} & \multirow{2}{*}{3} \\
\midrule
 \multirow{2}{*}{BERT-Attack} & Davis is \textcolor{blue}{often} \textcolor{blue}{enamoted} of her own \textcolor{blue}{generation} that she can not see how \textcolor{blue}{insuffoure} the \textcolor{blue}{queen} is. &\multirow{2}{*}{\textit{Failure}}& \multirow{2}{*}{27.8} & \multirow{2}{*}{0.16} & \multirow{2}{*}{2}\\
\midrule
\multirow{2}{*}{\method}  & Davis is so \textcolor{blue}{charmed} of her own cre\textcolor{red}{k}ation that she can\st{'t} see how \textcolor{blue}{indefensible} the character is. \textcolor{blue}{@kjdjq2.} & \multirow{2}{*}{\textit{Success}}& \multirow{2}{*}{14.6} & \multirow{2}{*}{0.59} & \multirow{2}{*}{4} \\
\bottomrule
\end{tabular}
\caption{Adversaries generated by \method and baselines in MR dataset. The replaced words are highlighted in \textcolor{blue}{blue}. \textit{Failure} indicates the adversary fails to attack the victim model and \textit{success} means the opposite.}
\label{tab:case_study}
\end{table*}

Table \ref{tab:training} displays adversarial training results of all datasets. Overall, after finetuned with both original training datasets and adversaries, victim model is more difficult to attack. Compared to original results, accuracy of all datasets is barely affected, while attack success rate meets an obvious decline. Meanwhile, attacking model with adversarial training leads to higher modification rate, further demonstrating adversarial training may help improve robustness of victim models.

\subsection{Against Defense} 
\begin{figure}[t!]
    \centering
\includegraphics[width=\linewidth]{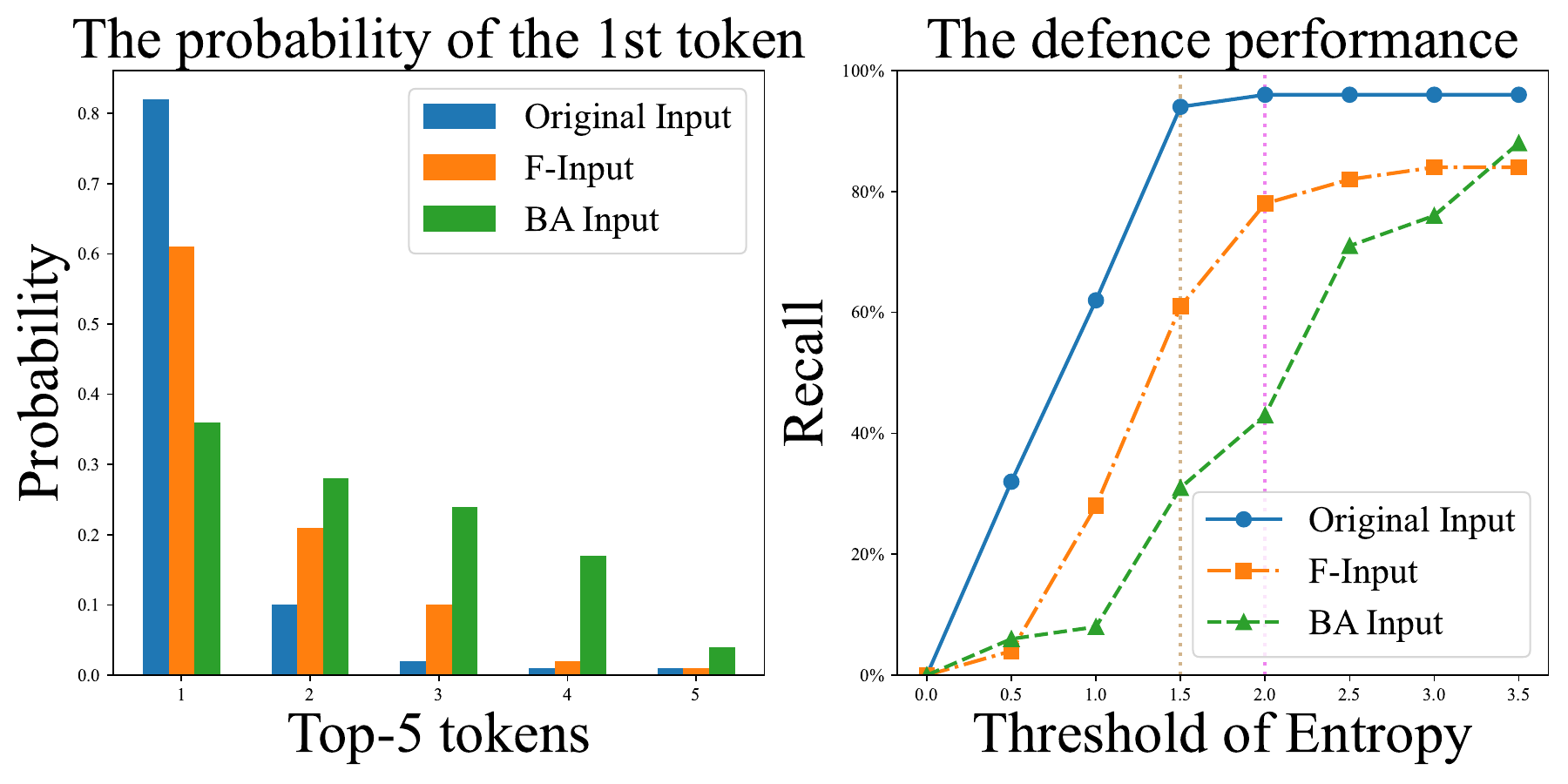}
    \caption{The probability of top-10 tokens in the first generated word in SA-LLaMA (a). The defense performance with various entropy thresholds (b)}
    \label{fig:defense}
\end{figure}

{Entropy threshold defense} \cite{yao2023llm} has been used to defend against the attack on LLMs recently. It employs the entropy of the first token prediction to refuse responding. Figure \ref{fig:defense} demonstrates the probability of top-10 tokens in the first generated word of LLaMA. It can be observed that the raw inputs usually generates the first token with low entropy (\textit{i.e.}, the probability of argmax token is much higher, and the probability of other tokens is much lower).
As shown in Figure \ref{fig:defense}, the adversarial samples from \method perform better than BERT-Attack with higher entropy. Attack samples generated through \method fare better against entropy-based filters compared to traditional text adversarial attack methods, indicating that the samples created by \method are harder to defend against.

\subsection{Case Study}
\label{app:case}
Table \ref{tab:case_study} shows adversaries produced by \method and the baselines. Overall, the performance of \method is significantly better than other methods. For this sample from the MR dataset, only TextFooler and \method successfully mislead the victim model, i.e., changing the prediction from \textit{negative} to \textit{positive}. However, TextFooler modifies twice as many words as the \method, demonstrating our work has found a more suitable modification path. Adversaries generated by TextFooler and BERT-Attack are failed samples due to low semantic similarity. BERT-Attack even generates an invalid word "\textit{enamoted}" due to its sub-word combination algorithm.  
We also ask crowd-workers to give a fluency evaluation. 

Results show \method obtains the highest score of 4 as the original sentence, while other adversaries are considered difficult to understand, indicating \method can generate more natural sentences.

\section{Conclusion}
In this paper, we examined the limitations of current adversarial attack methods, particularly their issues with transferability and efficiency when applied to Large Language Models (LLMs). To address these issues, we introduced \method, a new approach that uses an external overseer model to identify key sentence components and allows for parallel processing of adversarial substitutions. Our experiments on six benchmarks demonstrate that \method outperforms current methods, significantly improving both transferability and speed. Furthermore, the adversarial examples made by \method do not significantly affect the performance of model after it has been trained to resist attacks, thus strengthening its defenses. We believe that \method is a significant improvement in creating strong defenses against adversarial attacks on LLMs, with potential benefits for future research in this field.

\bibliography{acl_latex}

\appendix

\clearpage

\end{document}